\documentclass[conference]{IEEEtran}
\IEEEoverridecommandlockouts
\usepackage{cite}
\usepackage{amsmath,amssymb,amsfonts}
\usepackage{algorithmic}
\usepackage{graphicx}
\usepackage{textcomp}
\usepackage{xcolor}
\usepackage[colorlinks,linkcolor=red]{hyperref}
\usepackage{verbatim}
\usepackage{multirow}
\usepackage{xcolor}
\usepackage{amsmath}
\usepackage{booktabs}
\usepackage{hyperref}
\def\BibTeX{{\rm B\kern-.05em{\sc i\kern-.025em b}\kern-.08em
    T\kern-.1667em\lower.7ex\hbox{E}\kern-.125emX}}
\begin{document}

\title{H-CNN-ViT: A Hierarchical Gated Attention Multi-Branch Model for Bladder Cancer Recurrence Prediction}

\author{
\IEEEauthorblockN{
Xueyang Li\IEEEauthorrefmark{1},
Zongren Wang\IEEEauthorrefmark{2},
Yuliang Zhang\IEEEauthorrefmark{2},
Zixuan Pan\IEEEauthorrefmark{1},
Yu-Jen Chen\IEEEauthorrefmark{1},
Nishchal Sapkota\IEEEauthorrefmark{1},
Gelei Xu\IEEEauthorrefmark{1}, \\
Danny Z. Chen\IEEEauthorrefmark{1}, 
Yiyu Shi\IEEEauthorrefmark{1}\thanks{Corresponding author: Yiyu Shi (email: yshi4@nd.edu).}
}
\IEEEauthorblockA{\IEEEauthorrefmark{1}Computer Science and Engineering, University of Notre Dame, USA}
\IEEEauthorblockA{\IEEEauthorrefmark{2}Department of Urology, The First Affiliated Hospital, Sun Yat-sen University, China}
}

\maketitle

\begin{abstract}
Bladder cancer is one of the most prevalent malignancies worldwide, with a recurrence rate of up to 78\%, necessitating accurate post-operative monitoring for effective patient management. Multi-sequence contrast-enhanced MRI is commonly used for recurrence detection; however, interpreting these scans remains challenging, even for experienced radiologists, due to post-surgical alterations such as scarring, swelling, and tissue remodeling. AI-assisted diagnostic tools have shown promise in improving bladder cancer recurrence prediction, yet progress in this field is hindered by the lack of dedicated multi-sequence MRI datasets for recurrence assessment study. In this work, we first introduce a curated multi-sequence, multi-modal MRI dataset specifically designed for bladder cancer recurrence prediction, establishing a valuable benchmark for future research. We then propose H-CNN-ViT, a new Hierarchical Gated Attention Multi-Branch model that enables selective weighting of features from the global (ViT) and local (CNN) paths based on contextual demands, achieving a balanced and targeted feature fusion. Our multi-branch architecture processes each modality independently, ensuring that the unique properties of each imaging channel are optimally captured and integrated. Evaluated on our dataset, H-CNN-ViT achieves an AUC of 78.6\%, surpassing state-of-the-art models. Our model is publicly available at \href{https://github.com/XLIAaron/H-CNN-ViT}{https://github.com/XLIAaron/H-CNN-ViT}.
\end{abstract}

\section{Introduction}
\label{sec:intro}

Bladder cancer is the 10\textsuperscript{th} most common cancer worldwide, with over 570,000 new cases and 210,000 deaths annually~\cite{chen2024platelet}. The disease has a recurrence rate of up to 78\%~\cite{cxbladder_surveillance}, making precise diagnosis critical for patient survival and effective management. 
Following surgical intervention, patients commonly undergo follow-up Magnetic Resonance Imaging (MRI) scans to monitor for any residual cancerous tissues, as 
early detection and accurate identification of post-operative residual cancerous tissues can significantly reduce the recurrence risk and improve the 5-year survival rate, which currently stands at $\sim$77\% for localized bladder cancer cases~\cite{cancer_bladdersurvival}. 
However, post-operative imaging is highly complex due to the presence of scarring, swelling, and removal of visible cancerous regions, making accurate diagnoses difficult to even the most experienced radiologists. A highly accurate diagnostic method can greatly facilitate the diagnostic process, reducing unnecessary surgeries and treatments and improving the quality of life for patients.

Recent advances in deep learning (DL) have demonstrated transformative potential in medical imaging, particularly in complex cases where accurate diagnostic support is critical~\cite{jiang2023deep}. Among these advances, Vision Transformer (ViT)~\cite{dosovitskiy2020image} has gained recognition for its exceptional ability to capture global contextual information through self-attention mechanisms. Unlike Convolutional Neural Networks (CNNs)~\cite{wu2017introduction}, which are constrained by their dependence on localized receptive fields, ViT is adept at capturing information across extensive spatial regions, rendering it highly suitable for some medical imaging tasks in which critical features may be distributed throughout a scan. It has also demonstrated considerable success in a variety of medical imaging applications, including classification, segmentation, and detection~\cite{shamshad2023transformers}. However, as illustrated in~\cite{azad2023laplacian}, its emphasis on capturing global structural information often comes at the expense of precise localization, a limitation that poses challenges in capturing high-frequency components such as textures and edges. This deficiency is particularly problematic in tasks requiring detailed structural analysis, such as bladder cancer recurrence prediction, where texture and edge information is crucial for distinguishing residual cancerous tissue from post-operative scarring.

In related fields requiring a balance of global and local feature extraction, hybrid CNN-ViT models have been explored to leverage the complementary strengths of CNNs and ViT. For instance, Transformer-CNN Mixture (TCM) block~\cite{liu2023learned} combines CNNs with Swin Transformers~\cite{liu2021swin} in parallel to harness both feature types, achieving state-of-the-art results in image compression. Similarly, the Aggregate Enriched Features from CNN and Transformer (ACT) mechanism~\cite{yoo2023enriched} places CNN and Transformer blocks in parallel to enhance performance in image super-resolution tasks. Pooling-based Vision Transformer (PiT)~\cite{heo2021rethinking} enhances ViT’s performance by incorporating CNN-inspired spatial reduction, and ResNet-ViT~\cite{wahid2024hybrid} has demonstrated promising results in myocardial infarction detection. Despite these advancements, existing hybrid models that utilize CNN and ViT structures in parallel lack an effective strategy for fusing features extracted from each component, often resulting in redundant information and suboptimal feature integration. This limitation is especially pronounced in medical imaging applications, where both precise localization and global context are critical. Moreover, deep learning-based approaches for bladder cancer recurrence prediction remain underexplored, partly due to the lack of dedicated datasets for this complex task.

\begin{figure*}[t]
\centering
\includegraphics[width=\textwidth]{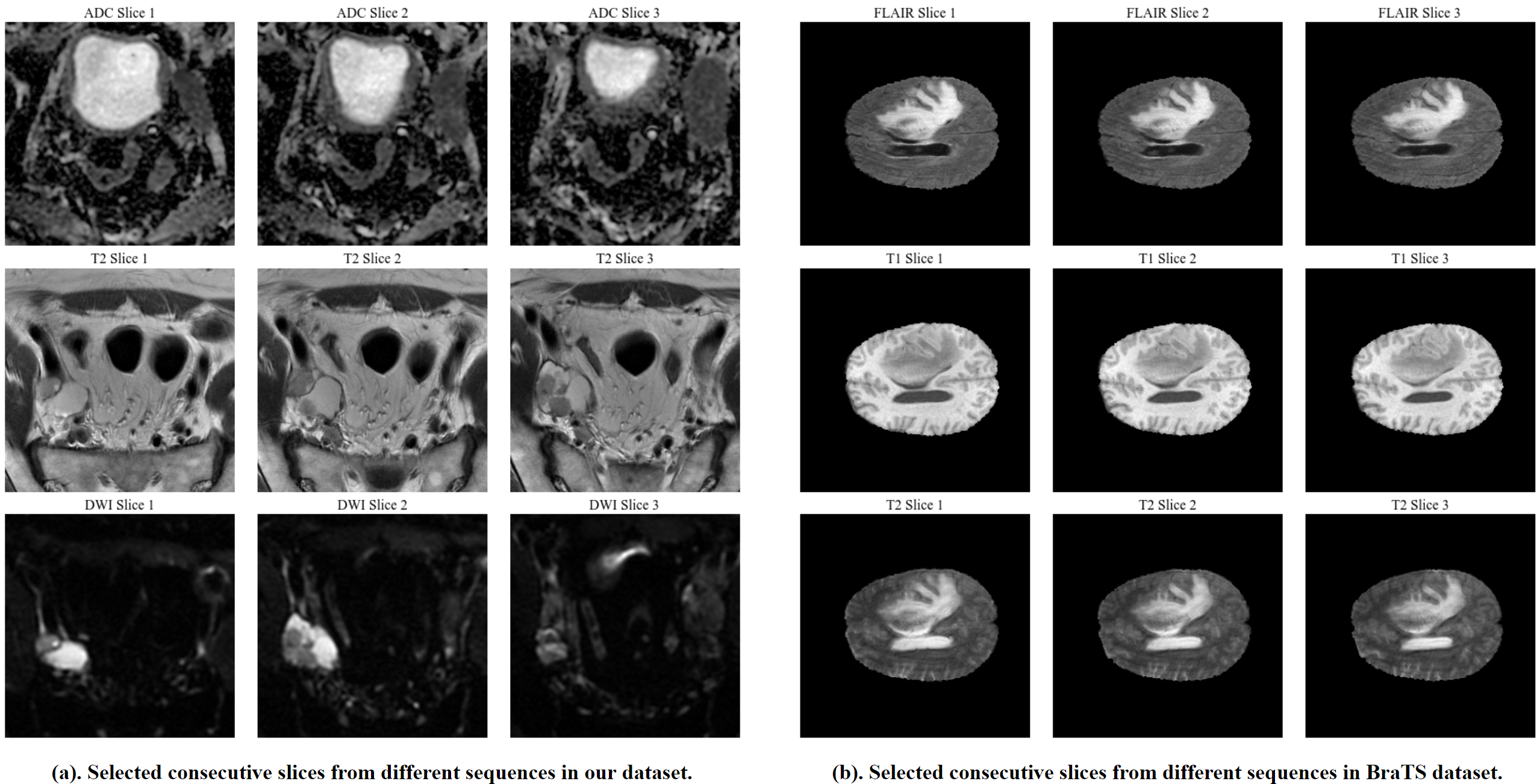}
\caption{Comparison of multi-sequence slices from our bladder cancer dataset and the BraTS  dataset~\cite{menze2014multimodal} to illustrate the distinct contrasts across the sequences in each dataset. (a) Consecutive slices from the ADC, T2, and DWI sequences in our bladder cancer dataset; (b) consecutive slices from the FLAIR, T1, and T2 sequences in the BraTS dataset.} 
\label{data1}
\end{figure*}

To bridge this gap, we introduce the first dataset specifically curated for bladder cancer recurrence analysis. Collected in-house from real-world clinical cases, our retrospective dataset presents unique challenges due to its high dimensionality and multi-sequence nature. Each patient's MRI scan includes three distinct sequences -- T2, ADC, and DWI -- each emphasizing unique tissue characteristics that must be comprehensively analyzed for accurate diagnosis (e.g., see Fig.~\ref{data1}). This forms a 4D dataset, in addition to the axial, sagittal, and coronal planes in normal 3D MRI scans. Unlike RGB images or even other multi-sequence medical datasets, \textbf{our dataset demands specialized handling.} For example, in common multi-sequence datasets like BraTS~\cite{menze2014multimodal}, sequences generally have the same slice count and share structural similarity, as illustrated in Fig.~\ref{data1}(b), making stacking a feasible approach. In contrast, our MRI bladder scans, as illustrated in Fig.~\ref{data1}(a), exhibit significant structural differences across the sequences, and further vary in slice count, making stacking impractical and registration infeasible. Besides, crucial clinical data -- such as patient sex, age, and preoperative tumor count -- must be incorporated to improve diagnostic accuracy. The complexity of managing \textbf{unregistered, multi-sequence, multi-modal} medical data poses a significant challenge to other state-of-the-art models, as they lack architectures specifically tailored for high-dimensional, heterogeneous medical datasets like ours.

Building upon this dataset, we propose a novel \textbf{H}ierarchical Gated Attention \textbf{CNN-ViT} (H-CNN-ViT) framework, designed to address the limitations of existing hybrid CNN-ViT models in handling high-dimensional, heterogeneous medical data. Unlike prior architectures that simply combine CNN and ViT in parallel, our model introduces a multi-branch design with hierarchical gated attention mechanisms to enable more effective feature refinement and fusion. Specifically, we design a Dual-Path Attention (DPA) block that integrates CNN and ViT pathways for extracting both local and global features. A first-level Gated Attention Module (GAM) fuses these features within each modality branch, while a second-level GAM aggregates the outputs across branches, ensuring a balanced integration of global context, fine-grained detail, and complementary modality information. We evaluate the proposed approach on our dataset and achieve an AUC (Area Under the Receiver Operating Characteristic Curve) of 78.6\%, surpassing strong baseline models and demonstrating the effectiveness of our method in addressing this challenging clinical task. We will release the dataset and code upon acceptance to facilitate further research in this domain. Therefore, our contributions can be summarized as follows:
\begin{itemize}
    \item We curate and will release a new multi-modal dataset specifically designed for the bladder cancer recurrence prediction task. The dataset includes unregistered 4D MRI volumes along with relevant clinical attributes, offering a valuable benchmark for this underexplored yet clinically important domain.
    \item We propose a novel framework that enhances existing CNN-ViT hybrid architectures through a hierarchical gated attention mechanism, enabling effective fusion of local and global features across multiple modalities in high-dimensional, heterogeneous medical data.
\end{itemize}

\section{Related Work}
\label{sec:Related_Work}
In this section, we 
review the current methods for dealing with the complexity of medical images, popular strategies for fusing CNN-ViT models, and gated attention mechanisms, which are fundamental topics on which our proposed H-CNN-ViT is based. Several
representative methods discussed in this section will be adopted in our experiments. 

\subsection{Complexities of Medical Images and Common Processing Strategies}
Medical images present unique challenges compared to natural images, mainly due to their high dimensionality and lack of direct alignment across modalities. Unlike natural images, which typically consist of a single, RGB-stacked channel, medical imaging modalities -- such as T2, ADC, and DWI sequences in MRI, or the unenhanced, portal venous, and arterial phases in contrast-enhanced computed tomography (CT) scans -- each captures different anatomical or functional properties, creating a multi-dimensional dataset that must be analyzed cohesively. Further, these modalities often have varying resolutions, slice counts, and spatial structures, making direct stacking or registration infeasible. As a result, this lack of alignment complicates feature extraction and requires specialized methods to fully leverage the complementary information in each modality.

To address these complexities of medical images, several strategies have been applied:
\begin{itemize}
    \item \textbf{Simple Stacking}: The most straightforward approach for handling different sequences or phases in medical images is simple stacking, in which the sequences are combined as separate channels, analogous to the 3 channels in RGB natural images. This approach enables the direct application of many mainstream DL models, such as X3D~\cite{feichtenhofer2020x3d}, ResNet~\cite{he2016deep}, EfficientNet~\cite{tan2019efficientnet}, and DenseNet~\cite{huang2017densely}. While this approach offers simplicity and may be feasible for datasets where images across sequences are relatively similar (e.g., BraTS~\cite{menze2014multimodal}), it fails to address the inherent misalignment in datasets with substantial differences across sequences, such as ours. As our experiments will show, this limitation hinders the ability to manage diverse slice counts and resolutions effectively, ultimately affecting the model performance.
    \item  \textbf{Multi-branch Structures}: An improved approach for handling multiple sequences in medical image datasets is the use of multi-branch architectures, where each modality is processed independently through separate network branches. This structure allows for modality-specific feature extraction, preserving each sequence’s unique characteristics before combining features at a later stage. Models employing this approach have shown considerable success, outperforming simple stacking methods that combine sequences into RGB-like channels~\cite{zhou2024fusing, li2023mpbd}. 
    \item \textbf{Conditional Models}: Recent methods, such as ConUNETR~\cite{sapkota2024conunetr}, employ conditional models in which all sequences are processed within a shared network with sequence-specific conditioning tokens. This approach seeks to create a unified representation by adaptively tuning the network to each sequence. But, as our experiments will demonstrate, this strategy is less effective than the multi-branch approach in handling the diverse characteristics of medical image sequences in our problem.
\end{itemize}

\subsection{CNN and ViT Feature Extraction Methods}
While there are other model types, such as Recurrent Neural Network (RNN) and Long Short-Term Memory (LSTM) network, for time-series data classification~\cite{sherstinsky2020fundamentals}, we focus on CNN and ViT based models due to their alignment with the spatial and contextual requirements of our task. CNN remains as one of the most widely-used models for image classification, leveraging convolutional layers to capture localized spatial features effectively. CNN’s hierarchical structure enables it to progressively detect complex patterns, making it highly effective for recognizing fine-grained details within an image~\cite{lecun1998gradient}. However, it is limited by the fixed receptive fields, which restrict the ability to capture global dependencies and broader contextual relationships. More recently, Vision Transformer (ViT) emerged as a powerful alternative, improving CNN by capturing global context through self-attention mechanisms. ViT treats image patches as tokens and models inter-patch relationships across the entire image, allowing for flexible, long-range dependency modeling~\cite{dosovitskiy2020image}. While ViT excels in capturing global features, it lacks the spatial localization precision provided by CNN, limiting its effectiveness in tasks requiring detailed local features.

Given their complementary strengths, a variety of methods have been proposed to combine CNN and ViT architectures, aiming to leverage CNN’s local feature extraction alongside ViT’s global context modeling. The Transformer-CNN Mixture (TCM) block~\cite{liu2023learned} and the Aggregate Enriched Features mechanism (ACT)~\cite{yoo2023enriched} arrange the CNN and ViT branches in parallel, yielding promising results in image compression and super-resolution tasks, respectively. 
ResNet-ViT~\cite{wahid2024hybrid} replaces the basic CNN with the more advanced ResNet structure, achieving notable success in myocardial infarction detection. However, these models lack a robust mechanism for effectively combining features extracted from the CNN and ViT components. For instance, ResNet-ViT attains fusion by simply concatenating features, while TCM and ACT apply a 1 $\times$ 1 convolution to the stacked features. These methods, however, fall short in achieving optimal feature integration, resulting in suboptimal performance due to limited interdependency modeling across CNN and ViT features.

\subsection{Gated Attention Mechanisms}
Gated attention mechanisms emerged as an effective tool for focusing selectively on relevant features within complex data, particularly in multi-modal tasks. As demonstrated in Gated Attention Networks (GaAN)~\cite{zhang2018gaan}, this mechanism excels in tasks involving complex spatiotemporal data by dynamically weighting features based on their relevance, enabling the model to prioritize critical information while reducing the impact of less informative features. The gated-attention-enhanced output $Y$ can be calculated as:
\begin{equation}
    Y = \sum_{i=1}^{N} {\alpha_i}\cdot{z_i},
    \label{eq:1}
\end{equation}
\vspace*{-0.15in}
\begin{equation}
    {\alpha_i} = \sigma(g(z_i)),
    \label{eq:2}
\end{equation}
where $\alpha_i$ is the attention score of branch $i$, $z_i$ represents the extracted features from branch $i$, $g$ represents the gating function, and $\sigma$ denotes activation function(s), with $i \in \{1,\ldots, N\}$ and $N$ being the total number of branches.

Gated attention mechanisms showed strong performance in processing sequential data~\cite{xue2020not} and high-dimensional medical data~\cite{schlemper2018attention}. But, as highlighted in Gated Attention Network (GA-Net)~\cite{xue2020not}, its potential for integration with Transformer architectures remains largely unexplored.

\begin{figure*}[t]
\centering
\includegraphics[width=0.8\textwidth]{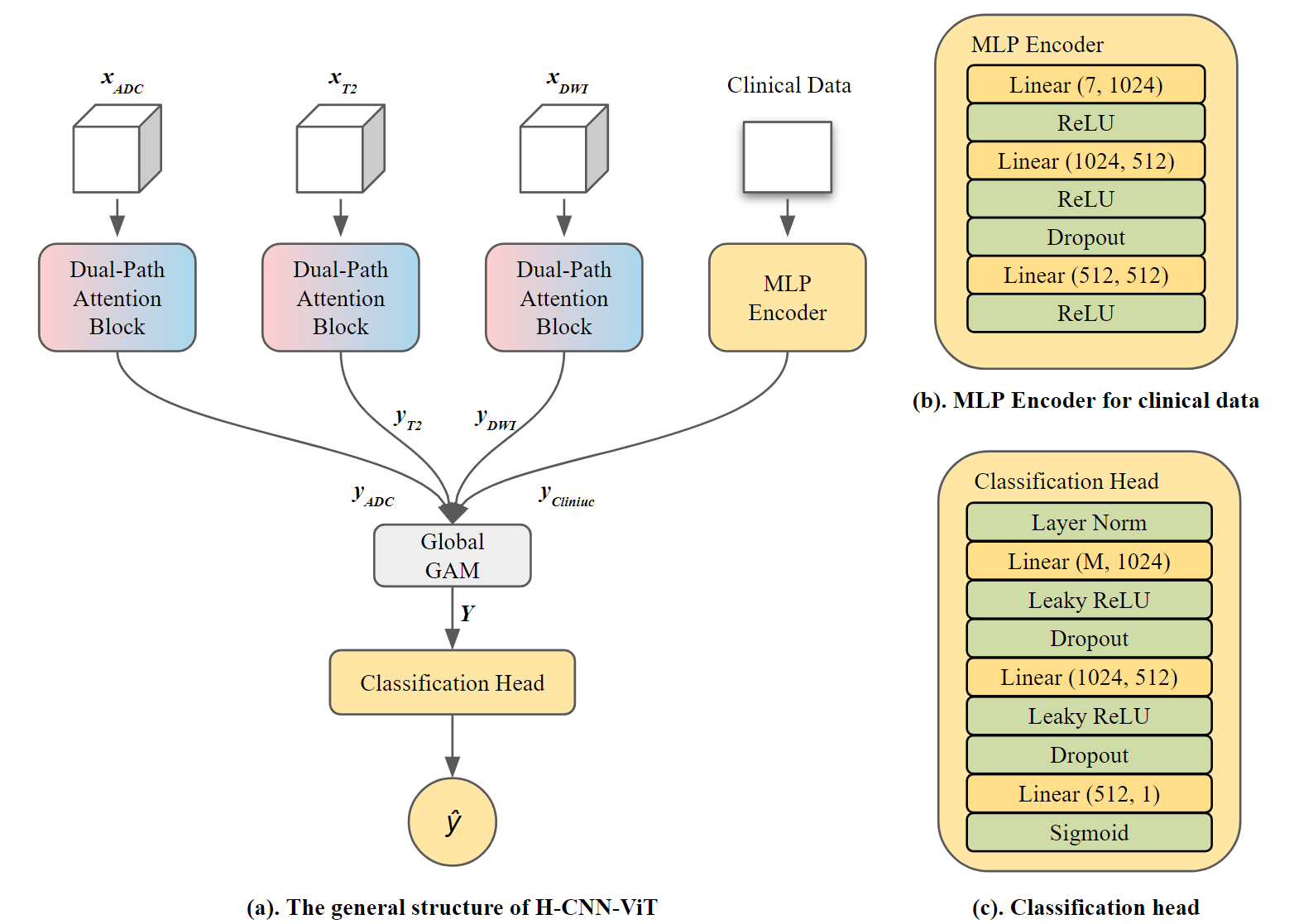}
\caption{(a) The overall framework of \textbf{H-CNN-ViT}, which includes three branches with Dual-Path Attention (DPA) blocks for different MRI sequences (ADC, T2, and DWI), an MLP (Multi-Layer Perceptron) Encoder for clinical data, and a Global Gated Attention Module (Global GAM) for integrating the extracted features ($y_{ADC}$, $y_{T2}$, $y_{DWI}$, and $y_{Clinic}$)  from each branch. The output from the Global GAM, $Y$, is passed to a classification head for final prediction. (b) The detailed architecture of the MLP Encoder for clinical data. (c) The detailed architecture of the classification head, which generates the final prediction. } 
\label{model1}
\end{figure*}

\begin{figure}[t]
\centering
\includegraphics[width=\columnwidth]{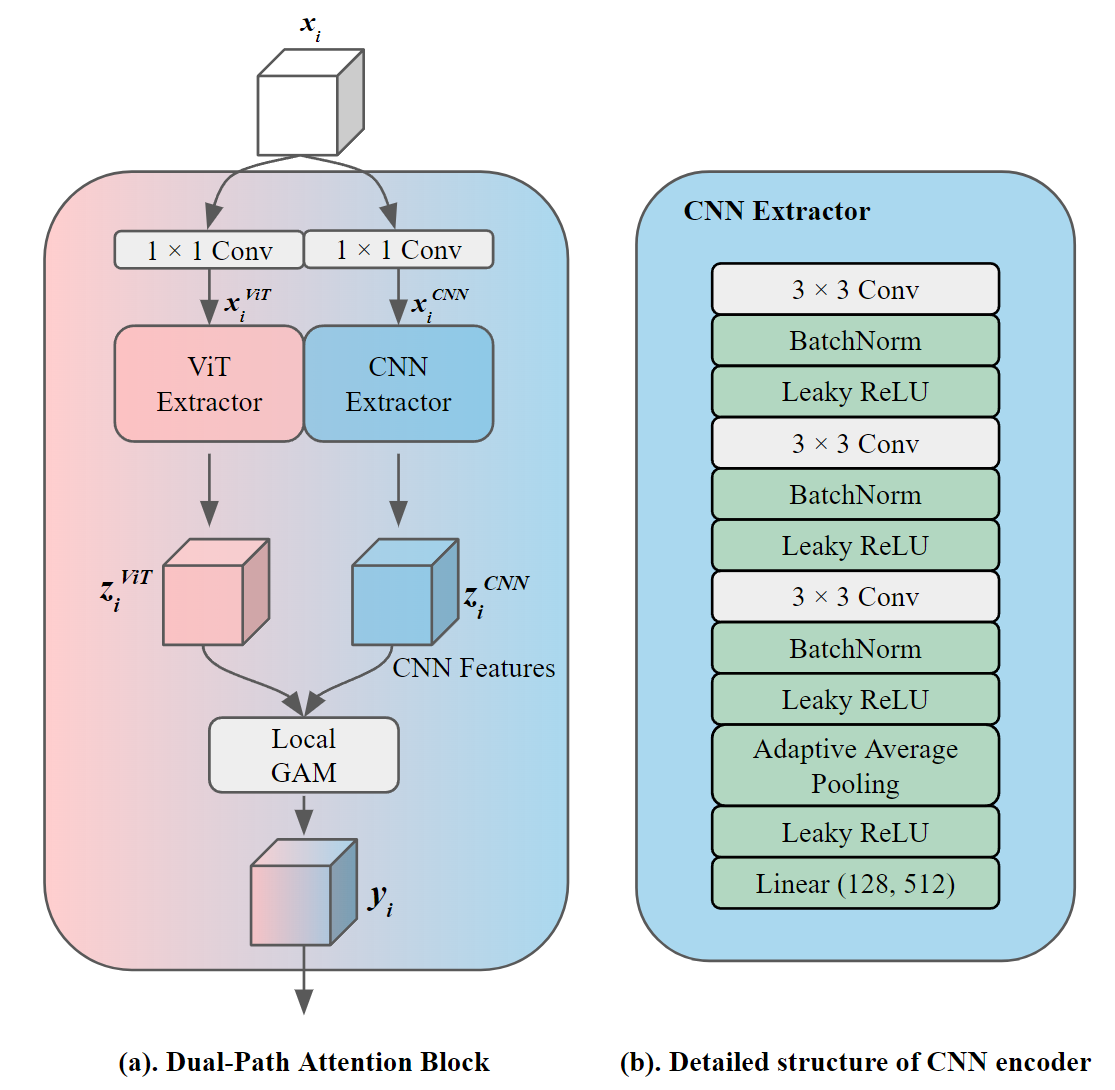}  
\caption{(a) The detailed structure of the Dual-Path Attention (DPA) block, which comprises two parallel paths: a ViT path and a CNN path. Each path includes a 1 $\times$ 1 convolutional layer for channel transformation and a feature extractor. A Local GAM is applied to fuse the outputs of both paths. (b) The detailed architecture of the CNN extractor, designed to capture localized features.}
\label{model2}
\end{figure}
\section{Dataset}
Our medical team collected a retrospective, multi-modality bladder cancer recurrence detection dataset from 346 patients during their follow-up clinic visits, typically within 3 to 6 months after surgery. Each patient has multi-sequence MRI scans (ADC, T2, and DWI), obtained during follow-up imaging, along with corresponding clinical data in tabular form. Tumor-related clinical attributes were derived from post-operative pathological evaluations of tumor specimens excised during the initial surgery, while demographic information (e.g., sex, age) was collected from patient records. The ground truth recurrence status was determined by pathologists based on pathological examination during the patient’s follow-up visit, serving as the gold standard for recurrence assessment. Each MRI sequence image is of size 280 $\times$ 280, with slice counts of the sequences varying from 13 to 60. Fig.~\ref{data1}(a) shows selected representative slices from each sequence in our dataset. As we mentioned earlier, our dataset presents distinct structural variations and different slice counts across the sequences, making it challenging to process with conventional stacking or registration techniques. For comparison, Fig.~\ref{data1}(b) illustrates the more structurally 
similar multi-sequence images from the BraTS 2018 dataset, in which the sequences generally have the same slice count and structural alignment, different only in contrast. The high level of heterogeneity in our dataset, combined with the inclusion of essential clinical attributes, underscores its \textbf{unique value} of developing more sophisticated diagnostic models for bladder cancer recurrence prediction.


The attributes of our clinical data are listed as follows:
\begin{itemize}
    \item \textbf{Age}: Patient age, ranging from 38 to 83 years.
    \item \textbf{Gender}: Either male or female.
    \item \textbf{Number of Hospitalizations}: The number of times the patient has been hospitalized, ranging from 1 to 42.
    \item \textbf{Tumor Size}: The maximum diameter of the tumor before surgery, ranging from 0.3 cm to 8.3 cm.
    \item \textbf{Single or Multiple Lesions}: A binary indicator (0 or 1), indicating whether the patient had a solitary tumor or multiple lesions before surgery.
    \item \textbf{T Stage}: An integer ranging from 0 to 5, indicating the size and extent of the primary tumor's growth, with 5 representing the most advanced stage.
    \item \textbf{Grade (High or Low)}: A binary indicator (0 or 1) representing the aggressiveness of the cancerous cells.
\end{itemize}
Of the 346 patients, 215 were classified as positive cases, indicating cancer recurrence, while 131 were negative cases, indicating no recurrence. Both the code and dataset will be released upon acceptance of this article.

\section{Method}

Fig.~\ref{model1}(a) illustrates the general structure of our H-CNN-ViT, which comprises four branches: three branches dedicated to processing different MRI sequences (ADC, T2, and DWI) via Dual-Path Attention (DPA) blocks, and one branch for encoding clinical data through a Multi-Layer Perceptron (MLP) Encoder. The features extracted from each DPA block -- denoted as \( y_{\text{ADC}} \), \( y_{\text{T2}} \), and \( y_{\text{DWI}} \) -- along with the clinical data features \( y_{\text{Clinic}} \) from the MLP Encoder, are subsequently passed into the Global Gated Attention Module (Global GAM), which connects to a classification head to generate the final prediction. The detailed structures of the MLP Encoder and the classification head are shown in Fig.~\ref{model1}(b) and Fig.~\ref{model1}(c), respectively. 

This multi-branch design allows for effective handling of multi-sequence data in medical imaging by enabling sequence-specific feature extraction that preserves the unique characteristics of each sequence. Additionally, the multi-branch structure offers flexibility to integrate multi-modality data. Note that while clinical data are used in our problem, this design can also incorporate other data types, such as text (e.g., by replacing the MLP Encoder with a Large Language Model (LLM) Encoder).

The following subsections provide an in-depth presentation of our core components, including the DPA block and Global GAM. We also highlight our hierarchical design of the Gated Attention Module, composed of both the Local and Global GAM components, by showing how it enhances the integration of sequence-specific and global features.

\subsection{Dual-Path Attention Block}
Fig.~\ref{model2}(a) presents the architecture of the Dual-Path Attention (DPA) block. The process begins with an input 3D MRI volume \( x_i \), where $i \in \{ADC, T2, DWI\}$ denotes an MRI scan sequence. Unlike the TCM block~\cite{liu2023learned} which employs a single unified \( 1 \times 1 \) convolutional layer for feature transformations across encoders, we introduce two distinct \( 1 \times 1 \) convolutional layers. This method enables independent channel transformations for each branch, ensuring that each encoder receives a unique feature representation optimized for its respective processing pathway. Additionally, this design allows for adjustable input and output dimensions in each \( 1 \times 1 \) convolution, enhancing flexibility for applications beyond single-channel MRI sequences (e.g., multi-channel natural images), without having to alter the model architecture. The outputs of two \( 1 \times 1 \) convolutional layers, denoted as $x_i^{ViT}$ and $x_i^{CNN}$, are further passed into the ViT extractor and CNN extractor. 
We utilize the original ViT encoder~\cite{dosovitskiy2020image} as the ViT extractor, as it outperforms the Swin Transformer encoder~\cite{liu2021swin} in global feature extraction, which will be demonstrated in the Experiments section. The architecture of the CNN extractor, shown in Fig.~\ref{model2}(b), consists primarily of three \( 3 \times 3 \) convolutional layers optimized for localized feature extraction, compensating for ViT’s lack of localization capability. Leaky ReLU is used in the CNN extractor to mitigate early overfitting in this deep network, as suggested in~\cite{xu2015empirical}. Following feature extraction, the ViT and CNN extractors produce outputs \( z_i^{\text{ViT}} \) and \( z_i^{\text{CNN}} \), which prioritize global and localized features, respectively. These outputs are then passed to the Local Gated Attention Module (Local GAM) for further refinement.
\subsubsection{Local Gated Attention Module}
\begin{figure}[t]
\centering
\includegraphics[width=\columnwidth]{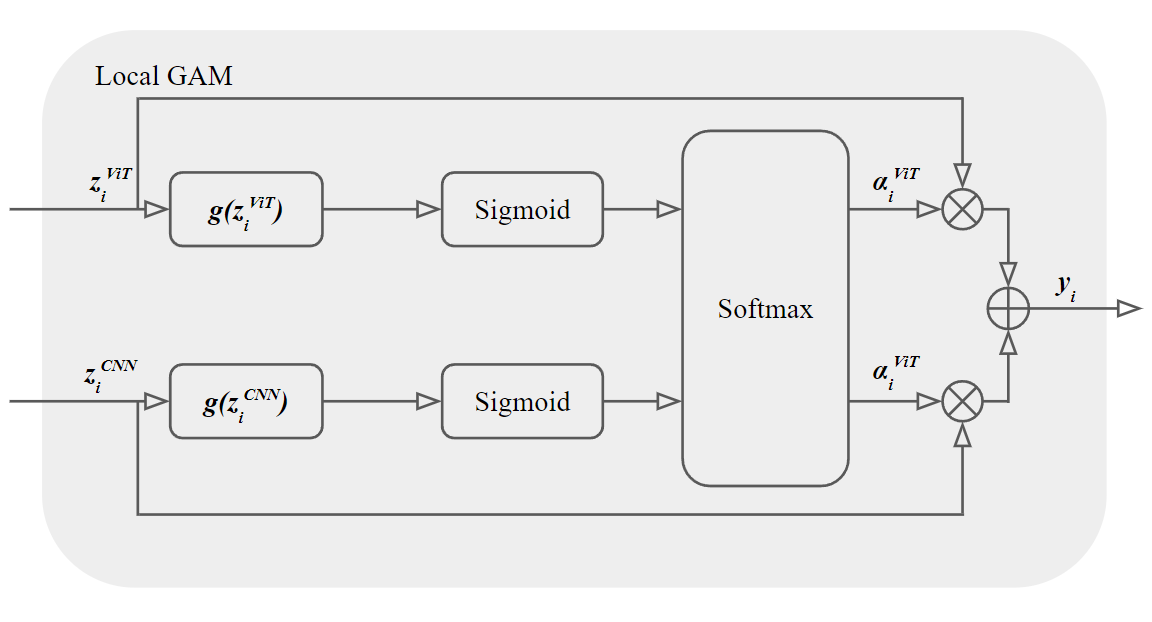}  
\caption{The Local Gated Attention Module for feature fusion within the Dual-Path Attention blocks.}
\label{local}
\end{figure}
Local GAM is employed to effectively fuse the extracted features from both extractors, balancing the contributions of global and localized information. Previous studies~\cite{zhang2018gaan, schlemper2018attention, xue2020not} have demonstrated the capability of gated attention in handling complex, high-dimensional data by selectively emphasizing relevant features while suppressing less informative ones. This dynamic feature weighting scheme makes gated attention an ideal choice for our model, as it enables a flexible integration of information from the ViT and CNN extractors. Fig.~\ref{local} illustrates the detailed structure of Local GAM. It begins with applying a trainable gating function, \( g(z_i^{\text{ViT}}) \) or \( g(z_i^{\text{CNN}}) \), to the features \( z_i^{\text{ViT}} \) or \( z_i^{\text{CNN}} \), iteratively refining this function during backpropagation to learn optimal weighting parameters. Each feature representation is then scaled by its respective gating function, dynamically adjusting the contributions of \( z_i^{\text{ViT}} \) and \( z_i^{\text{CNN}} \) to \( y_i \), the integrated gated-attention-enhanced output for branch \( i \). This process enables the model to prioritize the features based on their relevance to the task, achieving an adaptive balance between global and localized feature information.

In our design, the gating function \( g(z_i) \) is defined as a linear transformation:
\begin{equation}
    g(z_i) = W_i z_i + b,
    \label{eq:3}
\end{equation}
where \( W_i \) is a learned weight vector and \( b \) is a bias term. To bound the gate values between 0 and 1 -- allowing for partial, full, or no passage of feature information from each branch -- we apply a sigmoid activation to \( g(z_i) \). Further, to balance contributions across the extractors by ensuring the sum of all the gate values equal to 1, we normalize the gating function with a softmax function. Thus, the normalized attention scores for the ViT and CNN features within branch $i$, denoted as \( \alpha_i^{\text{ViT}} \) and \( \alpha_i^{\text{CNN}} \), can be calculated as:
\begin{equation}
    \alpha_i^{\text{ViT}}, \alpha_i^{\text{CNN}} = \text{softmax}([\sigma(g(z_i^{\text{ViT}})), \sigma(g(z_i^{\text{CNN}}))]),
    \label{eq:alpha1}
\end{equation}
where \( \sigma \) denotes the sigmoid function. This function maps the high-dimensional feature vector \( z_i \) to a scalar gate value \( \alpha_i \), which indicates the importance of that feature and serves as a soft filter. Building on Eqs.~(\ref{eq:1}) and~(\ref{eq:2}), the gated-attention-enhanced output of branch $i$, denoted as \( y_i \), can be calculated as:
\begin{align}
    y_i &= \alpha_i^{\text{ViT}} z_i^{\text{ViT}} + \alpha_i^{\text{CNN}} z_i^{\text{CNN}} \notag \\
    &= \frac{\exp(\sigma(g(z_i^{\text{ViT}})))}{\exp(\sigma(g(z_i^{\text{ViT}}))) + \exp(\sigma(g(z_i^{\text{CNN}})))} z_i^{\text{ViT}} \notag \\
    &\quad + \frac{\exp(\sigma(g(z_i^{\text{CNN}})))}{\exp(\sigma(g(z_i^{\text{ViT}}))) + \exp(\sigma(g(z_i^{\text{CNN}})))} z_i^{\text{CNN}}.
    \label{eq:5}
\end{align}
This mechanism enables Local GAM to adaptively adjust the importance of each feature representation, producing a fused representation \( y_i \) to capture the most relevant aspects of both the global and localized features within the branch. 

Building on the same mechanism, Global GAM applies similar attention principles across a broader scope, integrating information of multiple branches to create a cohesive representation.

\subsection{Global Gated Attention Module}

The Global Gated Attention Module (Global GAM) extends the functionality of Local GAMs by operating across four branches, instead of combining features from two extractors within a single branch. As shown in Fig.~\ref{model1}(a), each branch's output \( y_i \) is input to Global GAM. This module generates a final gated-attention-enhanced output, denoted as \( Y \), which integrates information from all the $N=4$ branches. Expanding on Eqs~(\ref{eq:1}) and~(\ref{eq:5}), \( Y \) is calculated as:
\begin{align}
    Y &= \sum_{i=1}^{N} {\beta_i}\cdot{y_i} 
    = \sum_{i=1}^{N} {\beta_i}(\alpha_i^{\text{ViT}} z_i^{\text{ViT}}  + \alpha_i^{\text{CNN}} z_i^{\text{CNN}}), 
    \label{eq:Y}
\end{align}
where \( \beta_i \) is the normalized attention score for branch \( i \), and \( i \) is one of the ADC, T2, DWI, and clinical data branches. The computation of \( \beta_i \) follows a similar design as that of the attention scores \( \alpha_i^{\text{ViT}} \) and \( \alpha_i^{\text{CNN}} \) within each branch. The gating function \( g \) is initialized as a linear transformation, and applies a sigmoid and a softmax activation, but here \( \beta_i \) takes the branch output \( y_i \) as input, rather than the extracted features \( z_i \) within the branch. Thus, \( \beta_i \) is computed as:
\begin{align}
    \beta_i &= \text{softmax}(\sigma(g(y_i))).
\end{align}

With this structure, we establish a \textbf{hierarchical gated attention} mechanism. The attention scores \( \alpha \) within each branch assign gating scores to the features extracted by the ViT and CNN pathways, balancing their contributions to the branch’s output \( y_i \). Subsequently, \( \beta \) computes gating scores across the branch outputs \( y_i \), further balancing each branch's contribution to the final prediction \( \hat{y} \). This dual-layer attention strategy enables the model to dynamically adjust the importance of both the local and global features in each branch while prioritizing the most informative modalities (ADC, T2, DWI, clinical data) for accurate prediction.

\section{Experiments and Results}
\begin{table*}[ht]
\centering
\caption{Performance comparison of different models on our dataset, evaluated in terms of AUC, Precision, and Recall. The last column shows the p-value for the AUC of each baseline model compared to our proposed H-CNN-ViT model, with p \textless{} 0.05 indicating statistically significant improvements by H-CNN-ViT over the baseline models.}
\begin{tabular}{llcccc}
\toprule
\textbf{Model Type} & \textbf{Model Name} & \textbf{AUC (\%)} & \textbf{Precision (\%)} & \textbf{Recall (\%)}   & \textbf{p-value}\\
\midrule
\multirow{3}{*}{CNN} & X3D-L~\cite{feichtenhofer2020x3d} & 73.9 $\pm$ 2.6 & 79.0 $\pm$ 5.4 & 86.1 $\pm$ 6.1 & 0.014\\
                     & ResNet18~\cite{he2016deep} & 75.4 $\pm$ 2.7 & 81.5 $\pm$ 4.5 & 91.0 $\pm$ 4.8 & 0.009\\
                     & DenseNet121~\cite{huang2017densely} & 73.0 $\pm$ 2.3 & 78.4 $\pm$ 5.9 & 88.4 $\pm$ 6.2 & 0.040\\ 
\midrule
\multirow{3}{*}{Transformer} & MB-ViT~\cite{zhou2024fusing} & 75.6 $\pm$ 3.1 & 80.1 $\pm$ 4.8 & 89.8 $\pm$ 5.5 & 0.032\\
                             & Swin Transformer~\cite{liu2021swin} & 71.3 $\pm$ 3.3 & 76.2 $\pm$ 5.7 & 88.7 $\pm$ 5.3 & 0.000\\
                             & ViT~\cite{dosovitskiy2020image} & 75.0 $\pm$ 2.6 & 79.2 $\pm$ 4.5 & 89.1 $\pm$ 5.3 & 0.035\\
\midrule
\multirow{3}{*}{Hybrid} & ResNet-ViT~\cite{wahid2024hybrid} & 76.0 $\pm$ 2.5 & 81.0 $\pm$ 5.6 & 90.3 $\pm$ 4.8 & 0.011\\
                        & TCM~\cite{liu2023learned} & 70.1 $\pm$ 3.1 & 78.0 $\pm$ 6.3 & 84.4 $\pm$ 6.0 & 0.002\\
                        & H-CNN-ViT (ours) & \textbf{78.6 $\pm$ 1.7}  & \textbf{82.0 $\pm$ 4.9} & \textbf{92.5 $\pm$ 4.5} & N/A\\
\bottomrule
\end{tabular}
\label{table:comparison}
\end{table*}

\begin{table*}[ht]
\centering
\caption{Ablation study on different components of H-CNN-ViT.}
\begin{tabular}{llccc}
\toprule
\textbf{Ablated Component} & \textbf{Model Variation} & \textbf{AUC (\%)} & \textbf{Precision (\%)} & \textbf{Recall (\%)} \\
\midrule
\multirow{3}{*}{GAM} & H-CNN-ViT w/o Local GAM & 77.8 $\pm$ 2.3 & 80.6 $\pm$ 5.0 & 91.6 $\pm$ 4.6 \\
                      & H-CNN-ViT w/o Global GAM & 77.2 $\pm$ 2.1 & 80.4 $\pm$ 5.2 & 91.3 $\pm$ 4.8 \\
                      & H-CNN-ViT w/o Any GAM & 76.9 $\pm$ 3.0 & 79.5 $\pm$ 5.8 & 90.8 $\pm$ 5.3 \\
\midrule
\multirow{2}{*}{CNN Encoder} & H-CNN-ViT w/ ResNet18 Extractor~\cite{he2016deep} & 78.2 $\pm$ 2.4 & 81.0 $\pm$ 5.1 & 91.9 $\pm$ 4.9 \\
                              & H-CNN-ViT w/ DenseNet121 Extractor~\cite{huang2017densely} & 76.1 $\pm$ 3.0 & 80.8 $\pm$ 5.5 & 90.1 $\pm$ 5.2 \\
\midrule
\multirow{2}{*}{Branch Configuration} & H-CNN-ViT w/ Single Branch & 75.3 $\pm$ 6.2 & 79.7 $\pm$ 5.6 & 90.5 $\pm$ 4.7 \\
                                      & H-CNN-ViT w/ Conditional Single Branch & 76.8 $\pm$ 5.5 & 80.1 $\pm$ 5.3 & 90.8 $\pm$ 4.9 \\
\midrule
Data Modality & H-CNN-ViT w/ MRI Images Only & 73.2 $\pm$ 5.3 & 79.3 $\pm$ 6.2 & 85.2 $\pm$ 4.4 \\
\midrule
Baseline & Full H-CNN-ViT (All Modules Enabled) & \textbf{78.6 $\pm$ 1.7} & \textbf{82.0 $\pm$ 4.9} & \textbf{92.5 $\pm$ 4.5}  \\
\bottomrule
\end{tabular}
\label{table:ablation}
\end{table*}

\subsection{Implementation Details}
To accommodate the varying number of slices across different MRI sequences, we employ Spline Interpolated Zoom (SIZ)~\cite{zunair2020uniformizing} to uniformly select 13 representative slices from 
each sequence. Every selected slice is downsampled to a 256 $\times$ 256 size, facilitating more efficient processing by the model. Standard data augmentation techniques are applied, including random rotation in the range of $-30$ to 30 degrees, as well as the mixup method~\cite{zhang2017mixup}, to enhance model robustness and prevent overfitting. For our limited-size dataset, we apply 5-fold cross-validation, with an 80-20 split between training and testing. We use the Adam optimizer with Binary Cross Entropy loss, a batch size of 8, and a learning rate of \(1 \times 10^{-4}\). Training spans a maximum of 400 epochs, with early stopping after 50 patience epochs. In the ViT extractor, the patch size is 16, frame patch size is 1, embedding dimension is 1024, and depth is 6. Dropout and embedding dropout are set at 0.2 and 0.1, respectively. Training on an NVIDIA A100 GPU with 80GB memory, each fold converges in $\sim$40 minutes, totaling $\sim$200 minutes for 5 folds. Training requires around 70GB of memory. A smaller batch size is feasible on GPUs with less memory.

For baseline selection, we consider three categories of models, and select representative architectures in each: (1) CNN-based models, including X3D-L~\cite{feichtenhofer2020x3d}, ResNet18~\cite{he2016deep}, and DenseNet121~\cite{huang2017densely}; (2) ViT-based models, including the original ViT~\cite{dosovitskiy2020image}, Swin Transformer~\cite{liu2021swin}, and Multi-Branch Vision Transformer (MB-ViT)~\cite{zhou2024fusing}; (3) hybrid models, which fuse CNN and ViT features, including TCM~\cite{liu2023learned} and ResNet-ViT~\cite{wahid2024hybrid}. For fair comparison, all models are trained on the same dataset using a 5-fold cross-validation scheme and incorporate both MRI and clinical data with a standardized extractor and MLP classification head, as detailed in the Method section. We use default configurations for each model except for TCM, which we adapted from its original 2D U-Net-like structure~\cite{ronneberger2015u} by re-implementing it in 3D, removing the decoder, and adding our MLP head. Due to the binary classification nature of this task, we mainly assess performance using AUC scores, supplemented by Precision and Recall scores. The classification threshold for Precision and Recall is set to 0.5.

\subsection{Results and Discussion}
\begin{figure}[t]
\centering
\includegraphics[width=\columnwidth]{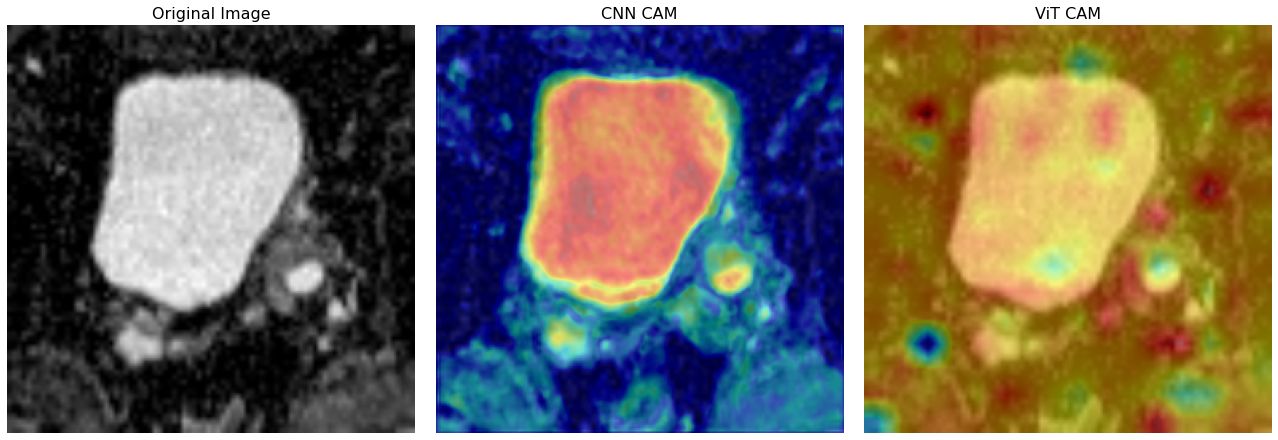}  
\caption{The leftmost image shows the original selected ADC slice used as input to the model. The middle image overlays the Class Activation Map (CAM)~\cite{zhou2016learning} derived from the third convolutional layer of the CNN extractor for the ADC modality. The rightmost image overlays the CAM derived from the attention mechanism in the second transformer block of the ViT extractor for the ADC modality.}
\label{CAM}
\end{figure}

Table~\ref{table:comparison} presents the AUC, Precision, and Recall scores for all the experimented models. As presented, our H-CNN-ViT outperforms all state-of-the-art models across the three metrics and achieves the lowest standard deviation in AUC scores, the primary metric for binary classification tasks, underscoring its superior stability. H-CNN-ViT consistently achieves a p-value of less than 0.05 compared to all baseline models, indicating statistically significant improvements. Note that some models, such as DenseNet121, demonstrate lower performance yet higher p-values due to overlapping performance distributions across folds. Additional result visualizations are provided in Fig.~\ref{CAM}. Note that all the models show higher Recall than Precision at the threshold of 0.5 due to our imbalanced dataset, which is beneficial in the context of bladder cancer recurrence prediction as higher Recall reduces the likelihood of missed diagnoses. The relatively low performance of TCM may be attributed to its original design for image compression tasks, and despite modifications, its architecture remains suboptimal for our classification task. Additionally, the improved performance of ResNet-ViT over the original ViT further supports the effectiveness of hybrid CNN-ViT architectures in handling complex, high-dimensional datasets such as ours.

As mentioned previously, we initially tested the Swin Transformer encoder as a possible feature extractor in the ViT path of H-CNN-ViT. However, as shown in Table~\ref{table:comparison}, ViT outperforms Swin Transformer by an average of 3.7\% in AUC, possibly due to our dataset’s ROI distribution. Swin Transformer’s localized attention may miss ROIs that are non-centrally located in images, while ViT’s global attention better accommodates broader ROIs. Further, ViT’s larger 16 $\times$ 16 patch size may capture features in larger spatial regions, aligning well with our dataset’s requirements.

\subsection{Ablation Study}

We conduct an ablation study, as presented in Table~\ref{table:ablation}, to evaluate each component’s impact on H-CNN-ViT’s performance, by selectively modifying elements such as GAM, CNN encoder, branch configuration, and data modality.

We first assess the effectiveness of GAM. As shown in Table~\ref{table:ablation}, removing Local GAM, Global GAM, or both GAM modules decreases the AUC score by 0.8\%, 1.4\%, and 1.7\%, respectively, alongside reductions in Precision and Recall. These results underscore the importance of our proposed Hierarchical Gated Attention Module in enhancing model performance. We further observe that replacing our current CNN extractor with a ResNet or DenseNet extractor also reduces performance. This may be attributed to the increased complexity of these extractors, which could lead to overfitting on our relatively small dataset.

\begin{figure}[t]
\centering
\includegraphics[width=0.8\columnwidth]{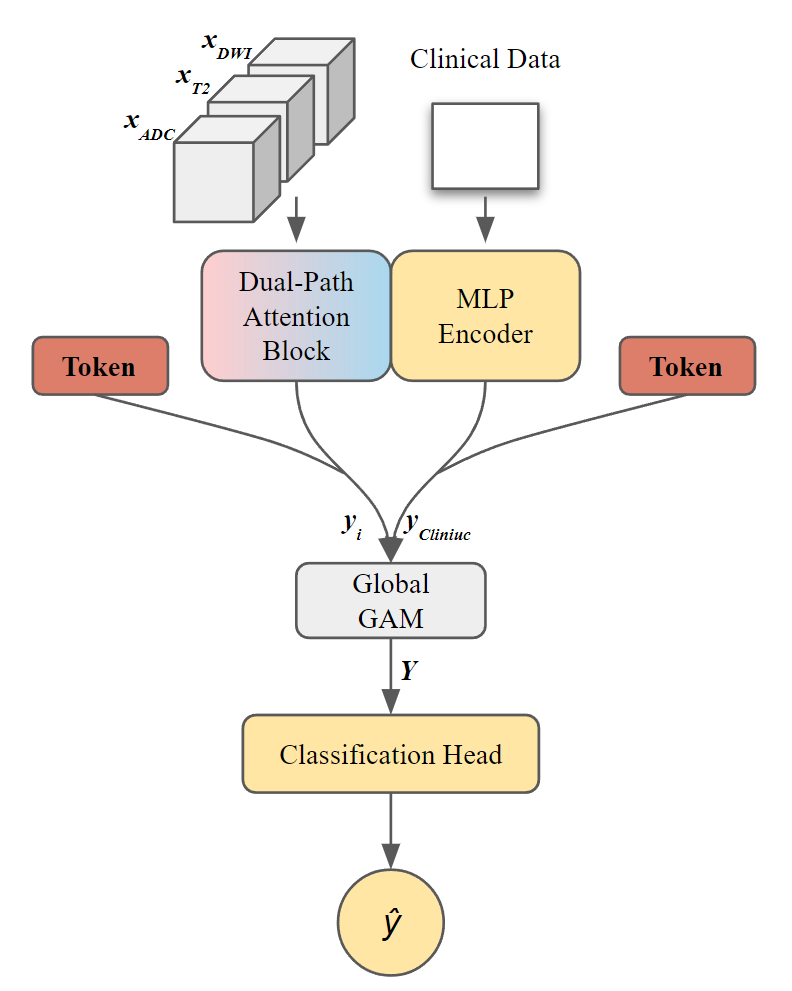}  
\caption{Illustration of the conditional model structure. MRI sequences (ADC, T2, and DWI) are processed sequentially through the Dual-Path Attention Block. After feature extraction, tokens indicating the respective MRI sequence are prepended to the features. Similarly, clinical data is passed through an MLP encoder, with a corresponding token denoting its type. The tokenized features from MRI sequences ($y_i$) and clinical data ($y_{\text{clinic}}$) are fused using the Global Gated Attention Module (Global GAM) to produce the combined representation ($Y$), which is subsequently passed to the classification head for the final prediction ($\hat{y}$).}
\label{condition}
\end{figure}

Additionally, we conduct an ablation study to verify the effectiveness of our multi-branch structure by testing a single-branch configuration that stacks multiple MRI sequences in an RGB-like fashion. This single-branch configuration results in a 3.3\% drop in AUC, supporting our view that simply stacking medical images as RGB-like channels can reduce performance due to suboptimal handling of sequence-specific features. Furthermore, removing Global GAM in the single-branch configuration likely contributes to this performance degradation as well. We also examine an alternative single-branch design that sequentially inputs MRI sequences along with a conditional token that is used to indicate the current sequence, as in~\cite{sapkota2024conunetr}. Although this design offers some improvement over the single-branch configuration, it still under-performs compared to the multi-branch structure, further validating our proposed approach. A detailed illustration of the conditional model is provided in Fig.~\ref{condition}.

Finally, the AUC score of 73.2\% achieved using image-only data highlights the importance of incorporating clinical data, demonstrating the effectiveness of our multi-modality approach in leveraging complementary medical information for improved prediction accuracy.

\section{Conclusions}
In this paper, we presented the first dedicated multi-sequence, multi-modal MRI dataset for bladder cancer recurrence prediction, addressing a critical gap in existing research. Building upon this dataset, we introduced H-CNN-ViT, a hierarchical gated attention model with a multi-branch architecture tailored for high-dimensional, heterogeneous medical data. The proposed Dual-Path Attention (DPA) block captures global and local features through parallel ViT and CNN pathways. To enhance feature integration, we introduced a two-level Gated Attention Module (GAM): a Local GAM for intra-branch fusion and a Global GAM for cross-branch aggregation. Our model demonstrated strong performance across AUC, Precision, and Recall, outperforming state-of-the-art baselines and highlighting the potential of our approach for improving diagnostic accuracy in complex clinical imaging tasks. Both the code and dataset will be released upon acceptance of this article.


\bibliographystyle{IEEEtran}
\bibliography{bibliography}

\end{document}